\documentclass{article}
\usepackage{ijcai25}

\usepackage{times}
\usepackage{soul}
\usepackage{url}
\usepackage[hidelinks]{hyperref}
\usepackage[utf8]{inputenc}
\usepackage[small]{caption}
\usepackage{graphicx}
\usepackage{amsmath}
\usepackage{amsthm}
\usepackage{booktabs}
\usepackage{algorithm}
\usepackage[switch]{lineno}

\usepackage{xspace}
\usepackage{cleveref}
\usepackage{tabularx}
\usepackage{multirow}
\usepackage{cleveref}
\usepackage[square]{natbib}
\usepackage{amssymb}
\usepackage{algpseudocode}
\usepackage{algorithm}
\usepackage{cleveref}
\usepackage{graphicx} 
\usepackage{caption}
\usepackage{tabularx}
\usepackage{booktabs}
\usepackage{multirow}
\usepackage[acronym]{glossaries}
\usepackage[nodisplayskipstretch]{setspace}
\usepackage[inline]{enumitem}
\usepackage{array}
\usepackage{subcaption}

\usepackage{thmtools} 
\usepackage{thm-restate}
\theoremstyle{plain}

\newcommand{\titlename}{{On the Effectiveness of Random Weights in Graph Neural Networks}}
\newcommand{\idenityemb}{Identity Embedding\xspace}
\newcommand{\fixedemb}{Fixed Random Embedding\xspace}
\newcommand{\dynamicemb}{Random Embedding\xspace}
\newcommand{\ours}{Learnable Embedding\xspace}

\newcommand{\fixed}{Fixed Random Diagonal\xspace}

\newcommand{\full}{Random Weights\xspace}
\newcommand{\fullfixed}{Fixed Random Weights\xspace}
\newcommand{\identity}{Identity Weights\xspace}
\newcommand{\pretrainedours}{\textsc{\method}\xspace}

\newcommand{\pretrainedemb}{Pretrained Embedding\xspace}

\newcommand{\method}{\textsc{RAP-GNN}\xspace}

\newcommand{\network}{f}
\newcommand{\classifier}{c_\theta}

\newcommand{\pretrainparam}{\phi}
\newcommand{\pretraininitlayer}{f^\text{pre}_{\pretrainparam}}
\newcommand{\pretrainf}{f^{\text{pre}}_\phi}
\newcommand{\pretrainc}{c^{\text{pre}}_{\pretrainparam'}}

\newcommand{\gnnw}{\mathbf{w}^{(l)}}

\newcommand{\gnnlayerl}{g_{\mathbf{w}^{(l)}}}

\usepackage{latexsym}

\newtheorem{observation}{Observation}
\newtheorem{definition}{Definition}

\Crefname{observation}{Observation}{Observations}

\title{\titlename}

\author{
Thu Bui$^1$\and
Carola-Bibiane Schönlieb$^2$\and
Bruno Ribeiro$^1$\and
Beatrice Bevilacqua$^1$\and
Moshe Eliasof$^2$\\
\affiliations
$^1$Department of Computer Science, Purdue University, West Lafayette, IN, USA\\
$^2$Department of Applied Mathematics, University of Cambridge, Cambridge, UK\\
}

\begin{document}

\maketitle

\begin{abstract}
    Graph Neural Networks (GNNs) have achieved remarkable success across diverse tasks on graph-structured data, primarily through the use of learned weights in message passing layers. In this paper, we demonstrate that random weights can be surprisingly effective, achieving performance comparable to end-to-end training counterparts, across various tasks and datasets. Specifically, we show that by replacing learnable weights with random weights, GNNs can retain strong predictive power, while significantly reducing training time by up to 6$\times$ and memory usage by up to 3$\times$. Moreover, the random weights combined with our construction yield random graph propagation operators, which we show to reduce the problem of feature rank collapse in GNNs. These understandings and empirical results highlight random weights as a lightweight and efficient alternative, offering a compelling perspective on the design and training of GNN architectures.
\end{abstract}

\section{Introduction}

\label{sec:intro}
Graph Neural Networks (GNNs) have emerged as a powerful tool for modeling and analyzing graph data ~\cite{bronstein2021geometric}. Traditionally, GNNs rely on learnable weights in their message-passing layers, which are optimized through end-to-end training to extract meaningful representations from graph data. While this paradigm has led to consistence performance, it also raises an intriguing question: \emph{Are learned weights always necessary for effective message passing?}

Previous studies have explored the role of randomization in neural network architectures. Random weights have been successfully applied in models like Extreme Learning Machines (ELMs)~\citep{ zhang2020graph,zhang2023semi} and Random Reservoir-Computing GNNs~\citep{gallicchio2020, Pasa2021MultiresolutionRG, huang2022are,navarin}, demonstrating that random and fixed projections can capture meaningful representations without extensive training.  Specifically, they rely on a random fixed set of weights, followed by a solution of a dynamic system based on these random weights. However, these approaches often lack a systematic understanding of why and how random weights contribute to effective learning. 

In this paper, \emph{we investigate the effectiveness of using random weights in GNNs as a replacement for learned parameters in message-passing layers}. Surprisingly, we find that GNNs with random weights can achieve performance comparable to their fully trained counterparts across a variety of graph tasks and datasets. This observation challenges the conventional reliance on end-to-end training to achieve strong predictive power in GNNs. 

To leverage the benefits of randomization, we propose \textbf{Ra}ndom \textbf{P}ropagation \textbf{GNN} (\method). Our method employs diagonal weight matrices, sampled on-the-fly at each forward pass, in combination with a simple pretrained node embedding layer. The diagonal structure preserves essential information about the dynamics of message passing while significantly reducing the number of parameters at each layer.

Beyond their effectiveness in terms of downstream performance, we show that random weights also offer computational advantages. By eliminating the need for gradient-based optimization of message-passing weights, \method reduces training time by up to 6$\times$ and memory consumption by up to 3$\times$. Moreover, we provide insights into why random weights are effective, by interpreting them as random graph propagation operators, which, combined with the diagonal weight structure, help to mitigate common GNN issues such as feature collapse during message passing.

Our findings suggest that random weights are not merely a computational shortcut but a principled design choice for scalable and effective GNN architectures, opening up new possibilities for resource-efficient graph learning systems.

The paper is organized as follows: In \Cref{sec:related}, we review works related to our approach.
\Cref{sec:method} introduces \method, highlighting its design and key advantages. We begin by outlining the architecture in \Cref{sec:architecture}, followed by an exploration of the critical issue of feature collapse in \Cref{sec:collapse}—a phenomenon where node embeddings lose diversity across layers. Next, in \Cref{sec:symmetry}, we discuss the symmetry-preserving capabilities of \method, showcasing its flexibility to function as either a permutation-sensitive or permutation-equivariant GNN, making it adaptable to a wide range of applications.
Finally, in \Cref{sec:exps}, we report an extensive set of experiments to demonstrate the effectiveness of \method.

\section{Related Work}
\label{sec:related}

\begin{figure*}[t]
    \centering
    \includegraphics[width=1.0\textwidth]{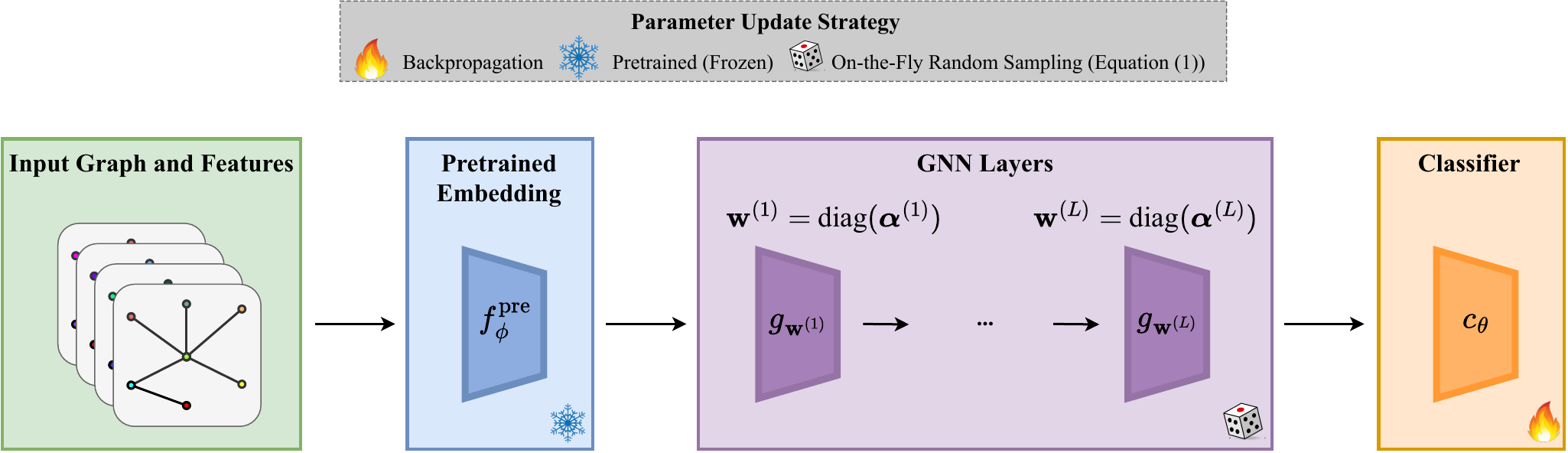} 
    \caption{Illustration of \textsc{\method}. The pretrained embedding \(\pretrainf\) is frozen, while the classifier \(\classifier\) is optimized. All GNN weights, $\mathbf{w}^{(l)}, \ l=1,\ldots,L$ are diagonal matrices, randomly sampled on-the-fly in each forward pass.
    }
    \label{fig:method}
\end{figure*}

The idea of using random weights as a substitute for learned parameters has been explored in various neural network paradigms. In the following, we discuss these approaches and provide additional references in \Cref{app:related}.

\paragraph{Random Reservoir Computing (RC).} 
Reservoir Computing (RC) is a computational framework introduced by~\citet{jaeger2001echo}, which leverages fixed, randomly sampled weights to parameterize a high-dimensional reservoir, i.e., a recurrent network that processes input data and projects it into a richer feature space. Traditional, i.e., non-neural, RC methods typically rely on recurrent forward passes until convergence or a predetermined number of iterations, and were shown to benefit graph tasks \citep{gallicchio2010}. Later, FDGNN~\citep{gallicchio2020} extended RC principles to GNNs for graph classification, where stability constraints are met in a recurrent setting. MRGNN~\citep{Pasa2021MultiresolutionRG} built on this by ``unrolling'' the recurrent layer, reducing time complexity while maintaining the benefits of reservoir dynamics. 
Despite these advancements, RC-based methods maintain fixed random weights, sampled only once at the beginning of training and used throughout. In contrast, our \textsc{\method} advocates for sampling new weights at each iteration. Moreover, RC methods are designed with specific architectures and graph tasks in mind (primarily for graph classification) whereas our \textsc{\method} represents a general framework that can be coupled with any GNN architecture to solve any graph task, enhancing its adaptability across various graph learning tasks. 

\paragraph{Extreme Learning Machines (ELM).} Another area of research has explored ELMs \citep{huang2006extreme}, which have recently been adapted for graph learning \citep{zhang2020graph, gonccalves2022extreme, zhang2023semi}. 
ELMs enable efficient training by using analytic, non-iterative learning techniques. However, in graph learning, these methods were shown to be constrained to two GNN layers and assume that the weights in the hidden layer are \textit{random and fixed} throughout the training process. 
This constraint hinders the ability of ELMs to leverage deep networks, which can be important for capturing the complex behaviors in the data.  In contrast, as shown in our experiments, our \textsc{\method} is effective across a variable number of layers and ensures that random weights are sampled at every forward pass, leading to more propagations seen during training time, and overall improved downstream performance. Additionally, by constraining \textsc{\method} to use \textit{fixed full (i.e., non diagonal) random weights}, a variant we denote as \fullfixed\ in our experiments, our approach can implement ELM. Thus, \textsc{\method}~can also be considered as a generalization of ELMs. 

\paragraph{Random Node Features (RNF).} Besides random graph models in terms of their weights, another research line suggests sampling random node features (RNF) to obtain theoretically improved performance \citep{murphy2019relational, abboudRNI, sato2021random, puny2020global}. In addition, the propagation of RNF has proven effective as a positional encoding technique \citep{eliasof2023graph} or within the normalization layer \citep{eliasof2024granola}. Differently, \textsc{\method} studies the value of random weights within the model, not random input features.

\section{Method}

\label{sec:method}
This section introduces our \method, a novel approach that utilizes diagonal random weights in the message passing layers of GNNs. In \method, the weights in each GNN layer are diagonal matrices, randomly sampled from a uniform distribution during each forward pass, thus eliminating the need for learning them.

We begin by detailing the architectural components in \Cref{sec:architecture}, providing an overview in \Cref{fig:method} and pseudocode in \Cref{app:algo}.
In \Cref{sec:collapse}, we introduce the concept of feature collapse, a significant challenge in GNNs, where node embeddings lose diversity and converge to nearly identical values as the number of layers increases. We demonstrate that \method effectively addresses this issue, ensuring stability even in deeper architectures. Finally, in \Cref{sec:symmetry}, we discuss the symmetry-preserving properties of \method, showing how it can be either permutation-sensitive or permutation-equivariant.

\paragraph{Notations.} In this paper we denote a graph by the tuple $G=(V,E)$, where $V$ is the set of $n$ nodes and $E$ is the set of $m$ edges. We further denote by $\mathbf{A} \in \{0,1\}^{n \times n}$ the adjacency matrix of the graph. We assume that the nodes are equipped with input node features of dimensionality $c$, denoted by $\mathbf{x} \in \mathbb{R}^{n \times c}$, and denote by $\mathbf{x}_i$ the $i$-th row of $\mathbf{x}$, which contains the features for node $i$. Throughout this paper, we will consider GNNs and denote the number of layers as $L$.

\subsection{\method} \label{sec:architecture}
Our \textsc{\method} consists of three key components: (i) a node embedding layer $\pretraininitlayer$, which is applied point-wise to the input node features of each node to obtain initial node embeddings $\mathbf{h}^{(0)} = \pretraininitlayer(\mathbf{x})$, (ii) a stack of $L$ GNN layers interleaved with non-linearities, where each layer $l = 1, \ldots, L$ is denoted by $\gnnlayerl$ and consists of one GNN layer with parameters $\mathbf{w}^{(l)}$ followed by a non-linear activation function, returning node embeddings, that is $\mathbf{h}^{(l)} = \gnnlayerl(\mathbf{h}^{(l-1)}, \mathbf{A})$, and (iii) a classifier $\classifier$, that takes the node embeddings and return a final prediction. The embedding layer $\pretraininitlayer: \mathbb{R}^{\text{c}} \rightarrow \mathbb{R}^{\text{d}}$ maps the $c$ input features into a hidden-dimensional space $d$, and is a multi-layer perceptron (MLP) or a single GNN layer. Each layer $\gnnlayerl: \mathbb{R}^{n \times \text{d}} \times \mathbb{R}^{n \times n}  \rightarrow \mathbb{R}^{n \times \text{d}}$ processes these embeddings along with the adjacency matrix, with weights $\mathbf{w}^{(l)}$, $l \in \{1, \dots, L\}$. In our \method, each $\mathbf{w}^{(l)} \in \mathbb{R}^{d \times d}$ is a diagonal matrix, where the diagonal values are randomly and uniformly sampled at each forward pass. This randomness eliminates the need to learn the GNN layer weights. If 
$\pretraininitlayer$ were learnable, backpropagation would still be required, even though the GNN layer weights would not be updated. However, we avoid this by pretraining 
$\pretraininitlayer$ separately, ensuring there is no backpropagation through the GNN. Finally, the classifier $\classifier: \mathbb{R}^{\text{d}} \rightarrow \mathbb{R}^{o}$ maps the processed representations (node or graph depending on the task) to the target output dimension $o$ to obtain the final prediction $\hat{y} = c_{\theta}(\mathbf{h}^{(L)})$, with only the classifier parameters $\theta$ being trained during the learning process. 
The model is trained in two phases:

\paragraph{(i) Pretraining Phase.} In \textsc{\method}, $\pretraininitlayer$ is responsible for processing input node features and mapping them into the desired hidden dimension $d$. Importantly, $\pretraininitlayer$ is trained in a pretraining phase to generate meaningful input representations. This pretraining step eliminates the need to train 
$\pretraininitlayer$ end-to-end during the main training phase, which would otherwise require backpropagation through the randomly sampled weights in the GNN layers.

The embedding layer $\pretraininitlayer$ is pretrained using a simple MLP or a single GNN layer, $\pretrainf = \pretrainc \circ \pretraininitlayer$, optimizing the downstream task. 
Once the embedding layer has been pretrained, we use $\phi$ during the main training phase, while keeping it frozen, allowing the model to learn only the classifier on the network.

\paragraph{(ii) Training Phase.} In the training phase, the key innovation is the introduction of   randomness in GNN layers. 
Namely, during each forward pass, a new random diagonal weight matrix \(\textbf{w}^{(l)}\) is sampled for each GNN layer $l$ as follows:
\begin{equation}
 \mathbf{w}^{(l)} = \mathrm{diag}(\boldsymbol{\alpha}^{(l)}),
\label{eq:rand_weight}
\end{equation}
where $\boldsymbol{\alpha}^{(l)} = [\alpha_1^{(l)}, \dots, \alpha_d^{(l)}]$ is a vector randomly sampled from a uniform distribution, with $d$ being the hidden dimension, and $l=1,\ldots, L$, with $L$ the total number of GNN layers in the network. 
\emph{Our motivation} for choosing the diagonal structure in \Cref{eq:rand_weight} over a full random matrix is threefold: (i) using a diagonal matrix for \(\textbf{w}^{(l)}\) replaces matrix multiplication with element-wise vector multiplication, reducing resource-intensive operations and accelerating both training and inference time, as demonstrated later in \Cref{sec:efficiency}; (ii) ensures that each feature is propagated independently by the corresponding $\alpha_i^{(l)}$, preventing random weights from overly mixing the features; and (iii) mitigates feature rank collapse, as demonstrated in \Cref{theorem:thm}. By constraining the values of $\mathbf{w}^{(l)}$ to the interval $[0,1]$ and using them as the weights of GNN layers such as GCN \citep{kipf2016semi}, we achieve diffusion-like propagation, similar to~\citet{eliasof2023improving}, which, in our case, because of the weights sampling approach, yields random graph propagation operators.

Note that, in this training phase, only the parameters $\theta$ of the classifier $\classifier$ are to be learned based on the loss function, while $\mathbf{w}^{(1)}, \dots, \mathbf{w}^{(L)}$ and $\phi$ are not learned. This eliminates the need for backpropagation through the GNN layers, significantly reducing computational costs, as illustrated in \Cref{fig:time_analyze_gnn}, and discussed in \Cref{sec:exps}.

\paragraph{(iii) Inference with \textsc{\method}.} During inference, the input is processed as a forward pass in the training phase, with random weights sampled from the uniform distribution.

\subsection{Reducing Feature Collapse with RAP-GNN} \label{sec:collapse}

Feature collapse is a critical issue in GNNs 
\citep{roth2024rank}, where node embeddings lose diversity and become nearly identical as the number of layers increases. This phenomenon hinders the ability of the network to learn and extract a variety of features to address downstream tasks, resulting in degraded performance. 
In this section, we consider feature collapse through the lens of feature matrix rank, defined below. Then, we theoretically demonstrate the impact of diagonal and full random matrices on node features rank, motivating the design choices in our \method. Lastly, we supplement our theoretical understanding with empirical evidence discussed below.

\noindent Formally, let $\bold{h}^{(l)} \in \mathbb{R}^{n \times d}$ denote the node embeddings at layer $l$ of a GNN.  
 We define the rank of $\boldsymbol{h}^{(l)}$ as follows:

\begin{definition}[Node Embedding Rank]
Assuming $n > d$, the maximal rank of the node embedding matrix $\boldsymbol{h}^{(l)}$ is $d$. Without loss of generality, let us denote the rank of $\boldsymbol{h}^{(l)}$ as $\text{rank}(\boldsymbol{h}^{(l)}) = d'$, where $1 \leq d' \leq d$.
\end{definition}

The rank of the embedding matrix is a useful indicator of the diversity of node embeddings. A higher rank suggests more diverse embeddings, whereas a lower rank indicates feature collapse, because more features are linearly dependent. To better understand this phenomenon, we analyze the rank dynamics under two types of random weight matrices: (i) full random matrices, and (ii) and diagonal random matrices, which we use in our \method. 

Specifically, we make the following observations, that highlight how the weight matrix structure affects the rank of the updated feature matrix, and why \method effectively mitigates feature collapse.

\begin{observation}[Rank Collapse with Full Matrix]
\label{obs:fullmatrix}
Consider a full (i.e., non diagonal) matrix $\mathbf{w}^{(l)} \in \mathbb{R}^{d \times d}$ whose entries are uniformly sampled between 0 and 1. Because in expectation, the value of each entry is 0.5, $\mathbb{E}[\mathbf{w}^{(l)}]$ can be expressed as:
\[
\mathbb{E}[\mathbf{w}^{(l)}] = 0.5 \cdot \mathbf{1}_{d \times d},
\]
where $\mathbf{1}_{d \times d}$ is a $d \times d$ matrix of ones. For simplicity, consider the update rule consisting only of the channel mixing operation defined as:
\[
\tilde{\boldsymbol{h}}^{(l)} = \boldsymbol{h}^{(l)} \mathbf{w}^{(l)}.
\]
In this case, we have:
\begin{align*}
\mathbb{E}[\tilde{\boldsymbol{h}}^{(l)}] =& \mathbb{E}[\boldsymbol{h}^{(l)} \mathbf{w}^{(l)}] = \boldsymbol{h}^{(l)} \mathbb{E}[\mathbf{w}^{(l)}] = \boldsymbol{h}^{(l)} (0.5 \cdot \mathbf{1}_{d \times d}).
\end{align*}
This implies that in expectation, each channel in $\tilde{\boldsymbol{h}}^{(l)}$ will be a linear combination of the channels in $\boldsymbol{h}^{(l)}$, with the same coefficient of 0.5. Consequently, all columns in $\tilde{\boldsymbol{h}}^{(l)}$ will be linearly dependent, resulting in $\mathbb{E}[\text{rank}(\tilde{\boldsymbol{h}}^{(l)})] = 1$.
\end{observation}

\Cref{obs:fullmatrix} demonstrates that using full random matrices can cause a significant reduction in rank, leading to rapid feature collapse as layers are stacked. In contrast, diagonal random matrices provide a more robust mechanism for preserving rank, as described next.

\begin{observation}[Rank Preservation with Diagonal Matrix in \method]
Consider the case of \emph{diagonal} random matrices $\mathbf{w}^{(l)}$, where the diagonal entries are uniformly sampled between 0 and 1. In expectation, each diagonal entry of $\mathbf{w}^{(l)}$ has a value of 0.5. Thus, $\mathbb{E}[\mathbf{w}^{(l)}]$ can be expressed as:
\[
\mathbb{E}[\mathbf{w}^{(l)}] = 0.5 \cdot \mathbf{I}_{d},
\]
where $\mathbf{I}_{d}$ is the $d \times d$ identity matrix. For simplicity, consider the update rule consisting only of the channel mixing operation defined as:
\[
\tilde{\boldsymbol{h}}^{(l)} = \boldsymbol{h}^{(l)} \mathbf{w}^{(l)}.
\]
In this case, each channel in $\boldsymbol{h}^{(l)}$ is scaled independently by the corresponding diagonal entry. In expectation, each scaling factor is 0.5. Thus, the rank of $\tilde{\boldsymbol{h}}$ remains unchanged, i.e.,
\[
\mathbb{E}[\text{rank}(\tilde{\boldsymbol{h}})]= \mathbb{E}[\text{rank}(\boldsymbol{h}^{(l)})] = d'.
\]
\label{theorem:thm}
\end{observation}

\Cref{obs:fullmatrix,theorem:thm} highlight the importance of the diagonal structure in random matrices used in \method. By preserving the rank of the feature matrix across layers, \method mitigates feature collapse and maintains diverse embeddings. 

The observations above establish that \method mitigates feature collapse especially when compared to GNNs with full random weights. In addition, we also observe in \Cref{fig:emb_var} that \emph{in practice \method exhibits  higher node embedding rank, and therefore less prone to feature collapse  than end-to-end trained GNNs}. Furthermore, we note that, although in principle, learnable weight matrices could adapt during training to mimic the behavior of \method (e.g., becoming approximately diagonal and bounded), we empirically show in \Cref{fig:gcn_weight_matrices} that the weight matrices in end-to-end trained GNNs do not approximate a diagonal structure after training.

\subsection{On the Symmetry Preservation in \textsc{\method}} \label{sec:symmetry}

In \textsc{\method}, each weight matrix  \(\mathbf{w}^{(l)}\) is sampled anew for every forward pass. This design ensures that the weights are consistent across all graphs within a batch but vary between batches. As a result, the output of \textsc{\method} is permutation-equivariant for each individual graph, but not across different graphs. Specifically, isomorphic nodes within a graph receive identical representations, while isomorphic nodes across different graphs may have distinct representations due to differing weights used to obtain representations between the graphs.

This design positions \textsc{\method} as a middle ground between permutation-equivariant GNNs, which consistently assign the same representation to isomorphic nodes regardless of their graph, and permutation-sensitive methods, such as those utilizing RNF, that differentiate all isomorphic nodes (with high probability). We believe that this hybrid approach facilitates more accurate predictions compared to RNF models, particularly in node classification tasks, as observed in our experiments. Since isomorphic nodes within a graph receive the same representation, the classifier does not need to map their distinct representations to the same prediction.

Notably, if we modify \textsc{\method} by sampling new \(\mathbf{w}^{(l)}\) matrices only once per epoch, instead of for every forward call, the method becomes full permutation-equivariant within that epoch. With this modification, \textsc{\method} consistently assigns the same representation to isomorphic nodes, regardless of their graph. Instead, if we incorporate RNF into \textsc{\method}, it inherits the permutation-sensitive properties from RNF. 
With these alternative designs, \textsc{\method} can act as a permutation-sensitive or permutation-equivariant GNN, or a hybrid of both. This flexibility enables users to tailor the model for various tasks that may require varying properties, broadening its applicability in different graph tasks.

\section{Experiments}
\label{sec:exps}
In this section we benchmark the performance of random weights in GNNs through our \method, and compare it with various baselines and methods, on a variety of datasets and graph learning tasks. Below, we elaborate on these baselines, following by outlining the research questions that our experiments seek to address.

\noindent\paragraph{Baselines.} To thoroughly evaluate the performance of \method, we consider the following relevant baselines:
\begin{enumerate}[label=(\roman*), leftmargin=*]
    \item \pretrainedemb: An embedding combined with a classifier is trained using an end-to-end method, intended for later use in \method and the other baselines discussed below. A \pretrainedemb can be either an MLP or a single-layer GNN, referred to as \(\pretraininitlayer\) in the Pretrained phase illustrated in \Cref{fig:method}. Additional details about the specific models used are provided in \Cref{app:hyperparameter_search}.
    \item End-to-End: A GNN backbone with learnable weights, that is trained in an end-to-end fashion, as is standard when training neural networks.
    \item \fixed: A variant using \textit{fixed} random diagonal weight matrices $\mathbf{w}^{(1)}, \ldots, \mathbf{w}^{(L)}$ in \Cref{eq:rand_weight}, that are sampled only once at network initialization, instead of sampling them on-the-fly. 
    \item \fullfixed: A variant using \textit{fixed} random \textit{full} matrices, that are sampled only once at network initialization, instead of sampling them on-the-fly. Notably, \fullfixed follows a similar principle to ELMs, where the weight matrices are sampled once and held fixed.
    \item \full: A variant using random \textit{full} matrices, that are sampled on-the-fly. 
\end{enumerate}

\noindent\paragraph{Research Questions.} Our goal is to address the following key research questions:

\begin{enumerate}[label=(Q\arabic*), leftmargin=*]
    \item How does \method compare to end-to-end training and other baselines of GNN architectures in terms of downstream performance?
    \item Is the choice of $\mathbf{w}^{(l)}, \ l=1,\ldots,L$ to be diagonal and sampled on-the-fly weight matrices superior to full or fixed random weights?
    \item How efficient  is our \textsc{\method} compared with end-to-end training? 
\end{enumerate}

We present our main results below, with additional experimental findings in \Cref{app:more_exps}. Particularly, \Cref{app:GNN_weights_variants} demonstrates that alternative choices of random weight matrix structures underperform compared to our design of \method, while \Cref{app:initial_layer} underscores the significant role of embedding pretraining. Furthermore, \method achieves accuracy comparable to end-to-end trained networks, while being significantly more efficient on large-scale datasets in \Cref{app:largegraphs_class}. We also show the robustness of \method to varying batch sizes, confirming its practical adaptability, in \Cref{app:batchsize}. Details on hyperparameter selection and dataset statistics are provided in \Cref{app:hyperparameter_search}.

\begin{table}[t]
\centering
\footnotesize
        \setlength{\tabcolsep}{3.5pt}
        \begin{tabular}{l ccc}
            \toprule
            Method $\downarrow$ / Dataset $\rightarrow$ & 
            \textsc{Cora} & 
            \textsc{CiteSeer} & 
            \textsc{PubMed} \\
            \midrule  
            \pretrainedemb & 56.55 $\pm$ 0.7 & 54.35 $\pm$ 0.5 & 72.00 $\pm$ 0.4 \\
            End-to-End & 81.50 $\pm$ 0.8 & {70.90 $\pm$ 0.7} & {79.00 $\pm$ 0.6} \\ 
            \fixed & 81.32 $\pm$ 0.8 & 70.12 $\pm$ 0.9 & 78.00 $\pm$ 0.9 \\ 
            \fullfixed & 58.70 $\pm$ 1.8 & 55.88 $\pm$ 1.2 & 71.52 $\pm$ 0.6 \\
            \full & 23.29 $\pm$ 9.2 & 45.85 $\pm$ 7.7 & 67.85 $\pm$ 1.7 \\
            \midrule
            \pretrainedours (Ours) & {82.42 $\pm$ 0.3} & 70.75 $\pm$ 0.3 & 78.94 $\pm$ 0.4 \\ 
            \bottomrule
        \end{tabular}
        \caption{Node classification accuracy (\%)$\uparrow$ using a GCN backbone and different training strategies. \textsc{\method} remains competitive or outperforms end-to-end training.}  
        \label{tab:GNN_weights_new}
\end{table}

\begin{table}[t]
    \centering
    \footnotesize
        \setlength{\tabcolsep}{3.5pt}
        \begin{tabular}{l ccc}
            \toprule
            Method $\downarrow$ / Dataset $\rightarrow$ & 
            \textsc{MUTAG} & 
            \textsc{PTC} & 
            \textsc{PROTEINS} \\
            \midrule  
            \textbf{\textsc{RC Methods}} \\
            $\,$ MRGNN 
            & N/A & 57.6 $\pm$ 10.0 & 75.8 $\pm$ 3.5 \\
            $\,$ U-GCN 
            & N/A & 62.6 $\pm$ 1.4 & 74.1 $\pm$ 0.9 \\ 
            $\,$ P-RGNN 
            & N/A & N/A & 71.1 $\pm$ 2.6 \\
            $\,$ FDGNN 
            & N/A & 64.3 $\pm$ 5.4 & {76.8 $\pm$ 2.9} \\ 
            \midrule
            \textbf{\textsc{Baselines}} \\
            $\,$ \pretrainedemb & 85.7 $\pm$ 7.9 & 59.0 $\pm$ 4.5 & 73.2 $\pm$ 3.8 \\
            $\,$ End-to-End  & 89.4 $\pm$ 5.6 & {64.6 $\pm$ 7.0} & 76.2 $\pm$ 2.8 \\
            $\,$ \fixed & 88.4 $\pm$ 7.4 & 63.3 $\pm$ 7.8 & 74.3 $\pm$ 3.4 \\
            $\,$ \fullfixed & 85.6 $\pm$ 7.8 & 62.8 $\pm$ 5.5 & 67.7 $\pm$ 5.0\\
            $\,$ \full & 74.0 $\pm$ 7.5 & 57.9 $\pm$ 8.4 & 60.3 $\pm$ 5.9 \\
            \midrule
            $\,$ \pretrainedours (Ours) & {90.4 $\pm$ 5.2} & 64.2 $\pm$ 9.8 & 75.9 $\pm$ 3.8 \\
            \bottomrule
        \end{tabular}
        \caption{Graph classification accuracy (\%)$\uparrow$ on TUDatasets. \textsc{\method} is competitive to end-to-end training and outperforms Reservoir Computing (RC) methods.}
        \label{tab:node}
\end{table}

\subsection{(A1) The Performance of \method}
\label{sec:gnn_weights}

We address Q1 by examining whether \method, that performs on-the-fly diagonal weights sampling, can serve as an effective alternative to an end-to-end training approach in terms of performance. To evaluate this, we compare the performance of \method against end-to-end trained networks and the aforementioned baselines across diverse graph tasks and datasets, leveraging various GNN backbones, detailed below.

\noindent \paragraph{Planetoid Node Classification.} \Cref{tab:GNN_weights_new} presents the performance comparison on the node classification task using the Cora \citep{mccallum2000automating}, CiteSeer \citep{sen2008collective}, and PubMed \citep{namata2012query} datasets. All experiments utilize a GCN backbone \citep{kipf2016semi} for the GNN layers with residual connections. The results demonstrate that \method also outperforms the end-To-end baseline on the Cora dataset and remains competitive on CiteSeer and PubMed. Notably, the results also highlight the benefits of on-the-fly sampling of diagonal weights, \(\mathbf{w}^{(l)}\), at each forward pass, because \method outperforms all other baselines involving random weights—\fixed, \fullfixed, and \full.

\noindent \paragraph{TUDatasets Graph Classification.} 
Here, we evaluate \textsc{\method} on a graph classification task. In our experiments, we use the popular GIN backbone \citep{xu2018how} on the TUDatasets \citep{morris2020tudataset}. In addition to the baselines evaluated in \Cref{tab:GNN_weights_new}, we include Reservoir Computing (RC) methods designed for graph classification and evaluated on the same datasets. 
The results in \Cref{tab:node} demonstrate that \method delivers competitive accuracy compared to end-to-end training.
Furthermore, \method, which employs randomly sampled diagonal weights on-the-fly, achieves accuracy comparable to Reservoir Computing methods such as MRGNN \citep{Pasa2021MultiresolutionRG}, U-GCN \citep{donghi2024investigating}, P-RGNN \citep{bianchi2022pyramidal}, and FDGNN \citep{gallicchio2020}, all of which rely on fixed random weights. These results underscore the effectiveness of \method.

\begin{table}[t]
\centering
\footnotesize
\setlength{\tabcolsep}{1.5pt}
\renewcommand{\arraystretch}{0.85} 
\begin{tabular}{l  ccc}
    \toprule
        \multirow{2}*{Method $\downarrow$ / Dataset $\rightarrow$} & \textsc{molesol} & \textsc{moltox21} & \textsc{molhiv} \\
        & \textsc{RMSE $\downarrow$} & \textsc{ROC-AUC $\uparrow$} &\textsc{ROC-AUC $\uparrow$} \\
        \midrule  
        \pretrainedemb & 1.235 $\pm$ 0.021 & 69.73 $\pm$ 0.39 & 70.97 $\pm$ 0.87 \\
        End-to-End & 1.173 $\pm$ 0.057 & {74.91 $\pm$ 0.51} & 75.58 $\pm$ 1.40 \\
        \fixed & 1.114 $\pm$ 0.060  & 71.73 $\pm$ 0.48 & 74.74 $\pm$ 0.93 \\
        \fullfixed & 1.122 $\pm$ 0.014 & 67.54 $\pm$ 0.51 & 71.82 $\pm$ 1.28 \\
        \full & 1.148 $\pm$ 0.029 & 67.61 $\pm$ 0.32 & 71.61 $\pm$ 1.32\\
        \midrule
        \pretrainedours (Ours) & {1.083 $\pm$ 0.053}  & 73.78 $\pm$ 0.51 & {76.04 $\pm$ 0.71} \\
        \bottomrule
\end{tabular}
\caption{Performance comparison of \textsc{\method} on graph classification tasks. The results show that \textsc{\method} achieves competitive performance to end-to-end trained GNNs. Additional results on the \textsc{molbace} dataset are provided in \Cref{tab:molbace}.}
\label{tab:no_molbace}
\end{table}

\begin{table}[t]
\footnotesize
\centering
    \setlength{\tabcolsep}{3.5pt}
    \renewcommand{\arraystretch}{0.85} 
    \begin{tabular}{lcc}
    \toprule
        \multirow{2}*{Method $\downarrow$ / Dataset $\rightarrow$} & \textsc{ogbn-proteins} & \textsc{ogbn-products}  \\
        &  \textsc{ROC-AUC $\uparrow$} & \textsc{Accuracy $\uparrow$} \\
    \midrule
    \pretrainedemb & 71.07 $\pm$ 0.36 & 72.83 $\pm$ 0.40 \\
    End-to-End & {77.29} $\pm$ 0.46 & 75.90 $\pm$ 0.31 \\
    \fixed & 76.84 $\pm$ 0.32 & 74.93 $\pm$ 0.78 \\
    \fullfixed & 73.71 $\pm$ 0.71 & 27.72 $\pm$ 0.79 \\
    \full & 70.46 $\pm$ 0.74 & 26.08 $\pm$ 0.82 \\
    \midrule
    \pretrainedours (Ours) & 77.01 $\pm$ 0.38 & 75.60 $\pm$ 0.88 \\ 
    \bottomrule
    \end{tabular}
\caption{Classification performance on ogbn-proteins and  ogbn-products using GCN~\citep{kipf2016semi}. \method is competitive to end-to-end training.}
\label{tab:ogbn-products} 
\end{table}

\noindent \paragraph{Graph-level Tasks on OGB.} 
Additionally, we evaluate \method on the OGB graph benchmark collection \citep{hu2020ogb} using the GINE backbone \citep{dwivedi2023benchmarking}. As shown in \Cref{tab:no_molbace}, \method demonstrates superior performance on the ogbg-molhiv dataset and remains competitive on the ogbg-moltox21 dataset compared to end-to-end. Notably, \method outperforms end-to-end on the regression task of the ogbg-molesol dataset, further highlighting its adaptability across various graph tasks. Results from the ogbg-molbace dataset, also part of the collection, are presented in \Cref{app:molbace}, where \method also outperforms end-to-end.

\begin{figure*}[t]
    \centering
    \includegraphics[width=1.0\textwidth]{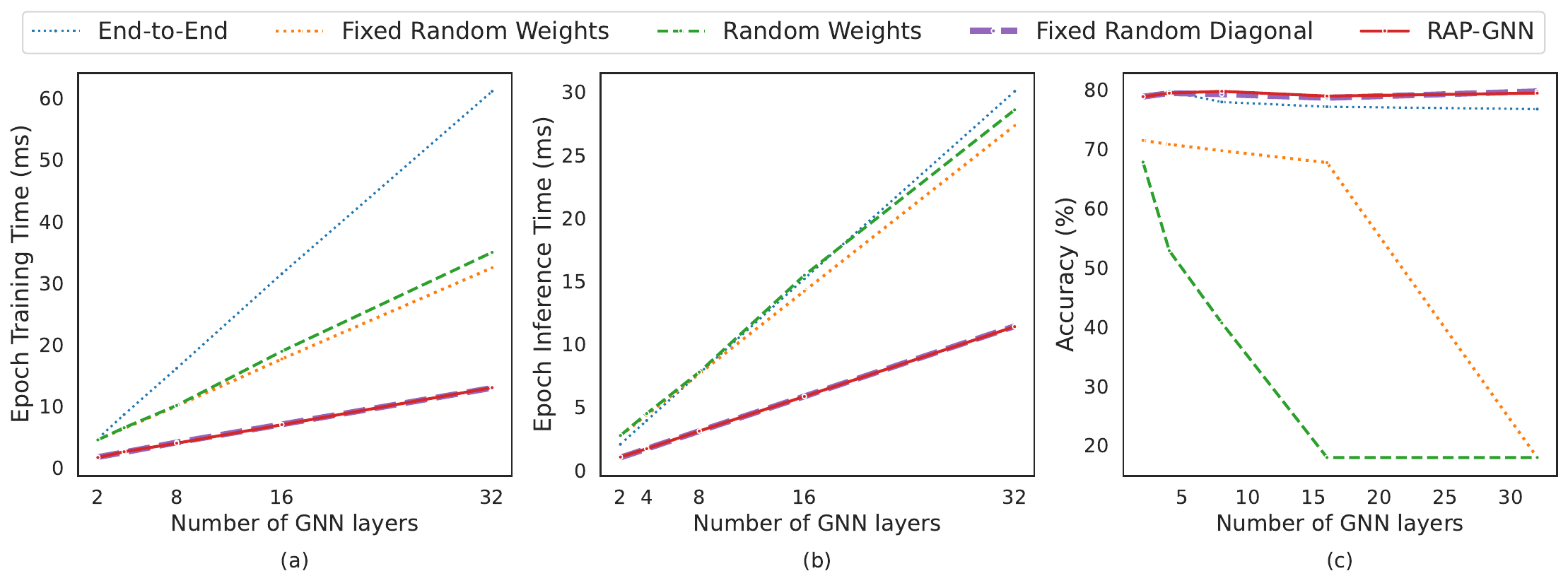} 
    \caption{
    Training time (a), inference time (b), and accuracy (c) for node classification on PubMed with GCN backbone. The runtimes gap between our \textsc{\method} and End-to-End widens as the number of layers increases, while obtaining superior accuracy than End-to-End. Accuracy results in (c) validate \Cref{theorem:thm}, showing \method maintains stability in deep architectures, unlike \full.
    }
    \label{fig:time_analyze_gnn}
\end{figure*}

\begin{figure}[ht!]
    \centering
    \includegraphics[width=0.48\textwidth]{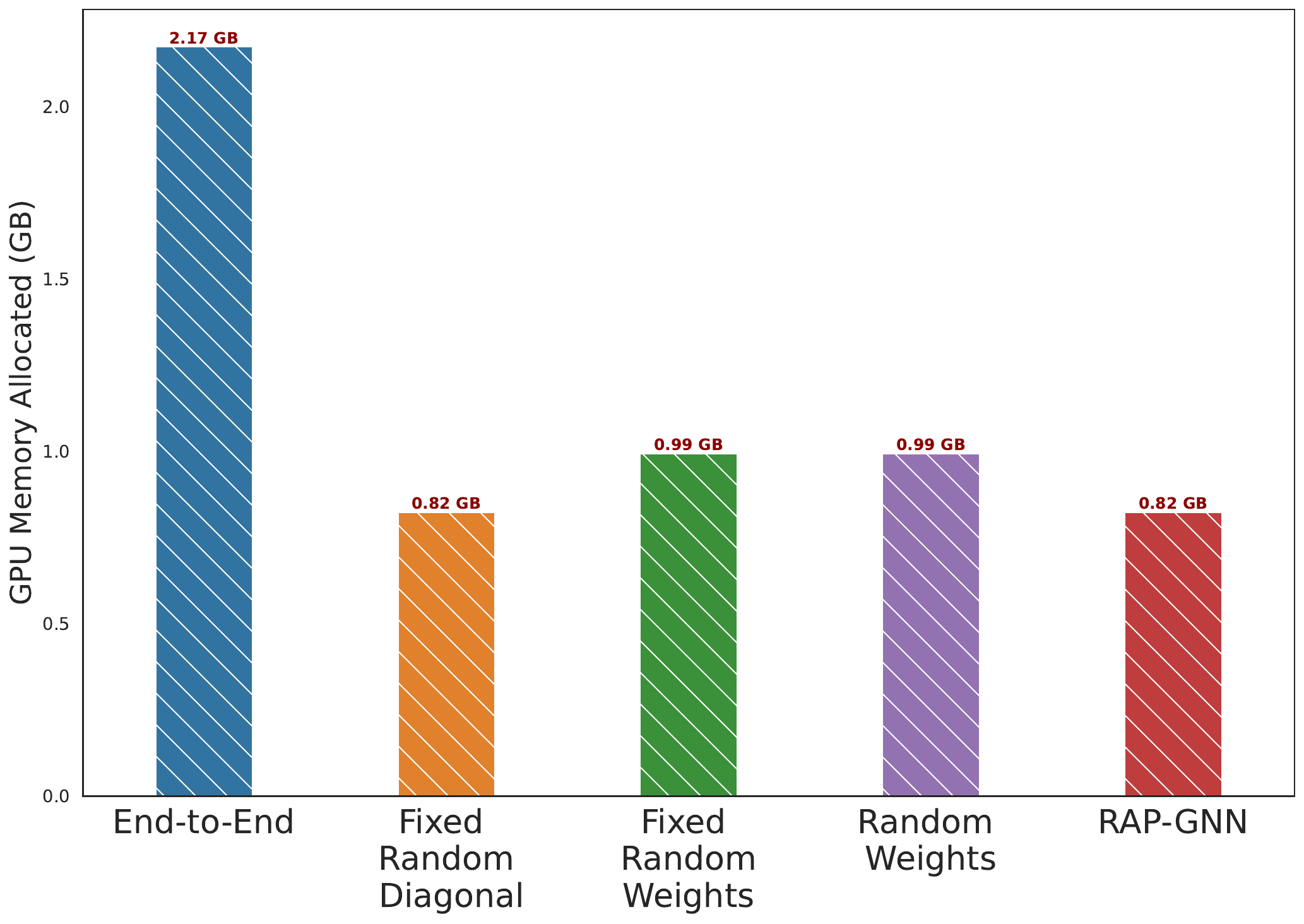}
    \caption{
    GPU memory usage comparison on on the PubMed dataset: \textsc{\method} requiring only a third of the memory compared to the End-to-End (GCN backbone).}
    \label{fig:gpu_memory_usage}
\end{figure}

\noindent \paragraph{Large-Scale Graph Node Classification.} To further demonstrate the applicability of \method, we now evaluate its performance on a node classification task using the large-scale {{ogbn-proteins}} and {ogbn-products} \citep{hu2020ogb} datasets.
The results presented in \Cref{tab:ogbn-products} (using GCN) indicate that \method delivers results on par with end-to-end training, while other baselines that employ random weights show a decline in performance, especially \full and \fullfixed on the ogbn-products dataset. A similar observation is made with GraphSage (\Cref{tab:ogbn-products-full} in \Cref{app:largegraphs_class}). These results underscore the effectiveness of \method in achieving competitive performance and its versatility across different GNN backbones.

\subsection{(A2) The Importance of On-The-Fly Sampling and Diagonal Weights}
\label{sec:a3}
In this section, we study the design of \method by examining its two key components: on-the-fly sampling and diagonal structured weights. First, we evaluate the significance of on-the-fly sampling by comparing \method to \fixed. Next, we explore the impact of the diagonal structure of \(\mathbf{w}^{(l)}\) by contrasting \method with two additional baselines: \full, which replaces diagonal weights in \Cref{eq:rand_weight} with full random matrices and sampling them on-the-fly, and \fullfixed, which also employs full random matrices but keeps them fixed during the training phase instead of sampling them on-the-fly.

\noindent \paragraph{The Importance of on-the-fly Sampling.}
The results in \Cref{tab:GNN_weights_new,tab:no_molbace,tab:ogbn-products} indicate that \fixed performs comparably to \method, though \method consistently achieves slightly better results across all evaluated datasets for node tasks. Furthermore, on graph classification datasets, as shown in \Cref{tab:node}, \method consistently outperforms \fixed. These findings are especially noteworthy given that RC methods share similarities with the fixed random weights approach used in \fixed, highlighting the advantage of on-the-fly sampling in delivering meaningful improvements over the fixed weight scheme.

\noindent \paragraph{The Importance of Diagonal Weights.} All the results in \Cref{tab:GNN_weights_new,tab:node,tab:no_molbace,tab:ogbn-products} show that the diagonal structure of $\mathbf{w}^{(l)}$ is beneficial for downstream performance. Namely, \pretrainedours consistently outperforms the variants using a full random matrix among different graph tasks in all datasets, whether coupled with on-the-fly sampling (\full) or fixed weights (\fullfixed).  
Specifically, both \full and \fullfixed exhibit a notable performance gap compared to both end-to-end and \pretrainedours. Moreover, using full-weight matrices achieves lower or similar performance of \pretrainedemb. {The experimental results in \Cref{fig:time_analyze_gnn}(c) underscore the significance of using diagonal weights and validate the findings of \Cref{theorem:thm}. They show that both \method and \fixed, which incorporate diagonal random weights, sustain stable performance even as architectures deepen. In contrast, \full and \fullfixed, using full random weights, experience a sharp performance decline as more layers are added, with \full exhibiting a rapid performance drop as more layers are added.}

\subsection{(A3) Memory and Time Consumption}
\label{sec:efficiency}

In previous experiments, we have shown that \textsc{\method} achieves accuracy comparable to end-to-end networks. Here, we shift the focus to empirically measure its efficiency. All the measurements discussed below are conducted on the PubMed dataset (19,717 nodes, 88,648 edges) using an NVIDIA GeForce RTX 4090 GPU.

\noindent \paragraph{Runtimes.} We measure the average runtime per epoch during training and inference, using networks with varying numbers of GNN layers, from 2 to 32, with 256 hidden channels.
\Cref{fig:time_analyze_gnn} shows the significant runtime efficiency of \textsc{\method} compared to end-to-end training on both training and inference. As the number of GNN layers increases, the efficiency gap between \textsc{\method} and end-to-end becomes more pronounced. For example, with 32  layers, \textsc{\method} reduces training time by approximately 48ms, offering a $6\times$ speedup. With 8  layers, \textsc{\method} shows a reduction of 12ms, a $3\times$ speedup, highlighting its scalability and efficiency. Although \fixed and \fullfixed also reduce training time by avoiding the need to train GNN layers and embedding layer, they remain slower than \method and \fixed. This is because \method and \fixed leverage diagonal weights during both training and evaluation, significantly reducing computations time.

\noindent \paragraph{GPU Memory Consumption.} Additionally, we measure the memory consumption of \method and end-to-end. To ensure a fair comparison, all networks are configured with 256 hidden dimensions and 32 layers.
In \Cref{fig:gpu_memory_usage}, the results highlight the resource efficiency of \textsc{\method}. {Although \full and \fullfixed show a reduction in GPU memory consumption compared to end-to-end, the decrease is more pronounced with \method and \fixed.} Both \textsc{\method} {and \fixed} use one-third the memory used by end-to-end.
As noted in \Cref{sec:method}, two key factors contribute to these gains. First, the design of \textsc{\method} eliminates the need for full backpropagation through the network, requiring gradient storage only for the final classifier layer. Second, the use of a diagonal matrix for \(\textbf{w}^{(l)}\) allows matrix multiplication to be replaced with element-wise vector multiplication, reducing the number of parameters and floating-point operations.

A similar trend of reduced training and evaluation time, along with lower GPU memory consumption, is observed for the large-scale datasets ogbn-products and ogbn-proteins in \Cref{app:largegraphs_class}. Combined with the performance effectiveness presented in \Cref{sec:gnn_weights}, these findings further highlight the applicability of \method for real-world applications, where efficient processing of large graphs is essential.

\section{Conclusion}
\label{sec:conclusion}

In this work, we present Random Propagation Graph Neural Networks (RAP-GNN), offering an alternative to traditional end-to-end training in GNNs by replacing learnable weights in message-passing layers with diagonal random weight matrices sampled on-the-fly. RAP-GNN reduces training time by up to 6× and memory usage by up to 3×, while maintaining competitive accuracy across diverse graph tasks and datasets. In addition, we show that our design choices allow RAP-GNN to mitigate feature collapse and maintain feature diversity, addressing challenges faced by deep GNN architectures. Our experiments demonstrate the scalability and efficiency of RAP-GNN, particularly on large-scale datasets, where it achieves strong predictive performance with significantly lower computational overhead compared to fully trained counterparts. This research advances the research frontier of resource-efficient graph learning and lays the groundwork for future exploration of scalable, lightweight GNN architectures for practical applications.

\bibliographystyle{named}
\bibliography{ref}

\clearpage
\newpage
\appendix

\section{Additional Details}
\label{app:algo}

In this section, we begin by exploring how weights are typically utilized in common GNN backbones.

We first present GNN backbones with \(\mathbf{w}^{(l)}\) and later show how they are random sampled on-the-fly. In our experiments, we use two popular backbones: GCN \citep{kipf2016semi} with residual connections, and GIN \citep{xu2018how}, defined as follows:

\textit{GCN.} The forward pass for the \(l\)-th GCN layer with a residual connection, where \(\sigma(\cdot)\) is a non-linear activation function (ReLU), is given by:
\begin{equation}
\mathbf{h}^{(l)} = \mathbf{h}^{(l-1)} + \sigma\Bigl(\tilde{\mathbf{A}} \mathbf{h}^{(l-1)} \mathbf{w}^{(l)}\Bigl),
\label{eq:gcn}
\end{equation}
where $\tilde{\mathbf{A}} = \mathbf{D}^{- \frac{1}{2}}\mathbf{A}\mathbf{D}^{- \frac{1}{2}}$, where $\mathbf{D}$ is the node degree-matrix.

\textit{GIN.} The forward pass for the \(l\)-th GIN layer (with $\epsilon=0$) reads:

\begin{equation}    
{
\mathbf{h}_v^{(l)} = \sigma\left(\mathbf{w}^{(l)}\Bigl(\mathbf{h}_v^{(l-1)} + \sum_{\forall u \in \mathcal{N}(v)}\mathbf{h}_u^{(l-1)} \Big)\right)
}
\label{eq:gin}
\end{equation}
where    \(\mathbf{h}_v^{(l)}\) is the embedding of node \(v\) at the \(l\)-th layer,   \(\mathcal{N}(v)\) denotes the set of neighbors of node \(v\), and \(\sigma(\cdot)\) is a non-linear activation function (ReLU). Notably, $\mathbf{w}^{(l)}$ can be extended to multiple weights $\mathbf{w}_j^{(l)}$ for GIN.

In \method, first we pretrain the node embedding $\pretraininitlayer$ as shown in \Cref{algo:rapgnn_pretrain}.
During the main training phase, \method samples $L$ random vectors $\boldsymbol{\alpha}^{(l)} \in [0, 1]^{d}$ ($l=1,\ldots,L$) and generates diagonal random GNN weight matrices $\mathbf{w}^{(l)}$ with $\boldsymbol{\alpha}^{(l)}$. Then \method uses these weights in the corresponding GNN layers (\Cref{eq:gcn,eq:gin}) to obtain the final representations and trains $\classifier$ on the downstream task. The pseudo-code for the training phase is provided in \Cref{algo:rapgnn_train}.

\section{Additional Related Work}
\label{app:related}

\paragraph{Random Strategies to perform Graph Data Augmentations.}
\textsc{GRAND} \citep{feng2020graph} enhances the robustness of GNNs by randomly dropping node features, either partially (via dropout) or entirely, making each node stochastically pass messages to its neighbors. This strategy increases robustness by making nodes less sensitive to specific neighborhoods and decouples feature propagation from transformation, enabling higher-order propagation without added complexity or over-smoothing. While \textsc{GRAND} improves performance in semi-supervised graph tasks by generating diverse augmented representations, it still requires learning all network parameters. In contrast, our method achieves enhanced performance by leveraging random propagation for message passing, completely bypassing end-to-end training.

\paragraph{Randomness in Recommendation Systems.}
A new line of research explores incorporating randomness into graph-based recommendation systems to enhance contrastive learning as well as more computational efficiency. For example, \citet{yu2022graph} propose adding noise directly to embeddings as an alternative to traditional graph augmentations in contrastive learning, introducing randomness into representation learning. In contrast, \method employs randomness in message propagation to boost performance.

\paragraph{Graph Lottery-Ticket Hypothesis.} 
The Graph Lottery-Ticket hypothesis, introduced in recent years \citep{Chen2021AUL,wang2023searching}, suggests that it is possible to identify sparse sub-networks that perform as well as end-to-end trained models. However, this approach requires pretraining the entire network to discover these sub-networks, which remains computationally intensive. Recent work expands this idea by identifying sub-networks within randomly weighted networks. \citet{ramanujan2020s} apply this to vision tasks, showing that pruning can match the performance of end-to-end optimized networks. Nevertheless, their method demands over-parameterized networks before pruning, increasing memory and computational requirements. \citet{chijiwa2021pruning} address this by re-randomizing weights during subsequent pruning iterations, reducing over-parameterization. More recently, \citet{huang2022you} adapted pruning mask optimization for graph tasks. 
Unlike these methods, which focus on identifying efficient sub-networks without modifying weight values from randomly initialized networks, our approach seeks to match or surpass the competitive performance of end-to-end trained models by random propagation.

\begin{algorithm}[t]
\caption{Pretraining  $\pretraininitlayer$.}
\label{algo:rapgnn_pretrain}
\begin{algorithmic}[1]
\State Initialize learning rate $\eta^\text{pre}$ and maximal number of epochs $\text{MaxPreEpochs}$ for the pretraining process.
\For{$i = 1$ to $\text{MaxPreEpochs}$}
    \For{$\mathbf{x}$ in $\text{Data}$}
    \State $\hat{y} = \pretrainc \circ \pretraininitlayer(\mathbf{x})$
    \State Compute the downstream task loss between the ground-truth $y$ and $\hat{y}$.
    \State Update $\pretrainc$ and $\pretraininitlayer$ parameters using a gradient-descent method (Adam) with a learning rate $\eta^\text{pre}$. 
    \EndFor
\EndFor
\end{algorithmic}
\end{algorithm}

\begin{algorithm}[t]
\caption{Training $\classifier$. 
}
\label{algo:rapgnn_train}
\begin{algorithmic}[1]
\State Initialize learning rate $\eta$ and max epochs $\text{MaxEpochs}$ for the training process.
\For{$i = 1$ to $\text{MaxEpochs}$}
    \For{$\mathbf{x}$ in $\text{Data}$}
    \State Sample $L$ random vectors $\boldsymbol{\alpha}^{(l)} \in [0, 1]^{d}$ \Comment{\(l=1,\ldots,L\)}
    \State Generate diagonal random GNN weight matrices $\mathbf{w}^{(l)}$ with $\boldsymbol{\alpha}^{(l)}$ \Comment{\Cref{eq:rand_weight}}
    \State Compute model output $\hat{y} = \network(\mathbf{x}, \mathbf{A})$
    \State Update parameters $\theta$ in classifier $\classifier$ using a gradient-descent method (Adam) with learning rate $\eta$. 
    \EndFor
\EndFor
\end{algorithmic}
\end{algorithm}

{\section{Rank Collapse}
\label{app:proofThm1}

\begin{figure*}[t]
    \centering 
    \includegraphics[width=0.6\textwidth]{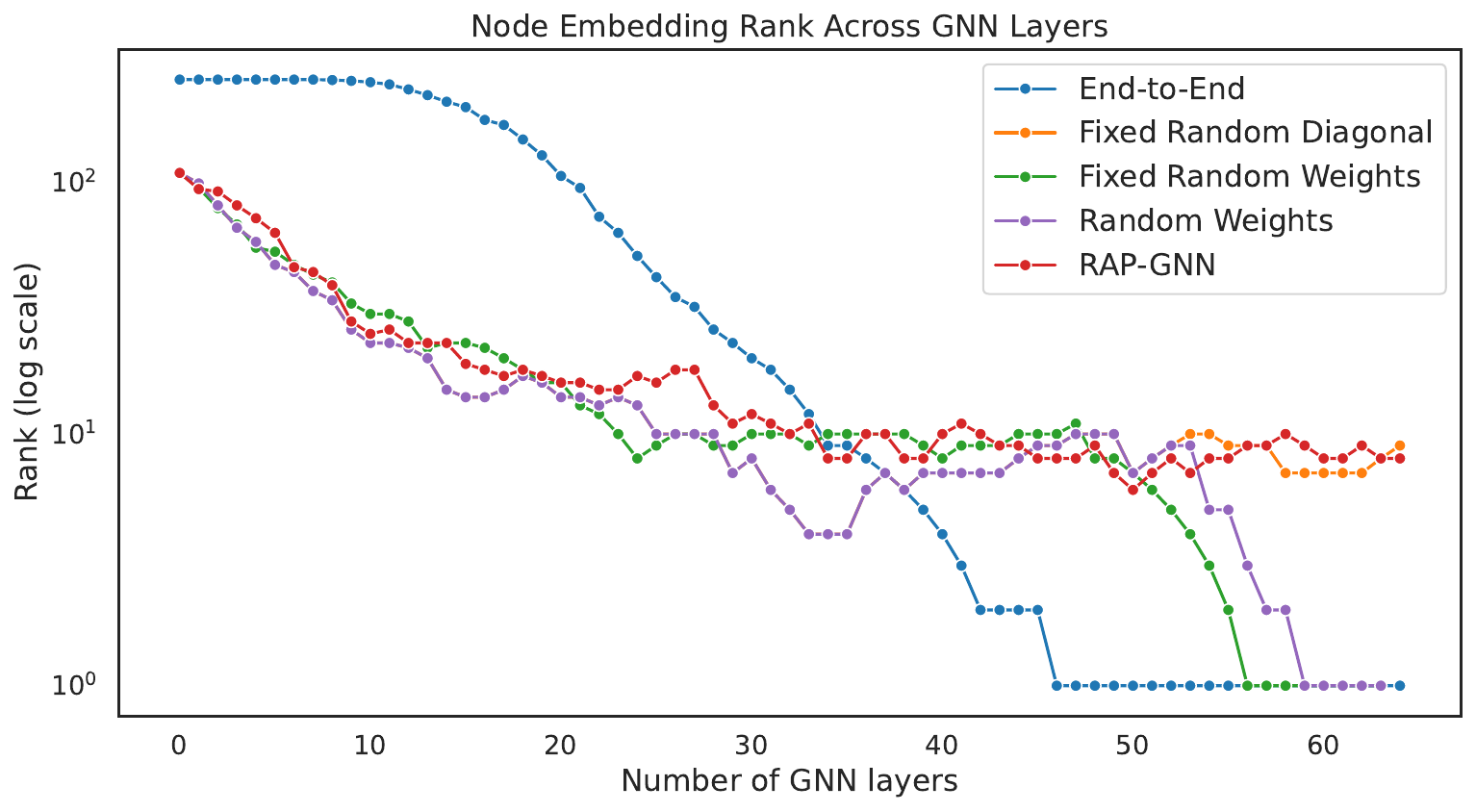} 
    \caption{Comparison of the mean of $\text{Var}\big(\bold{h}^{(l)}\big)$ (in log scale) across layers for End-to-End  and \method on the PubMed dataset using the same GCN architecture with residual connections with $L =64$ and $d=256$. \method increases embedding variance with higher rate compared to End-to-End as the number of layers grows, consistent with the theoretical result in \Cref{theorem:thm}.}
    \label{fig:emb_var}
\end{figure*}

\begin{figure*}[t]
    \centering
    \begin{subfigure}[t]{0.48\textwidth}
        \centering
        \includegraphics[width=\textwidth]{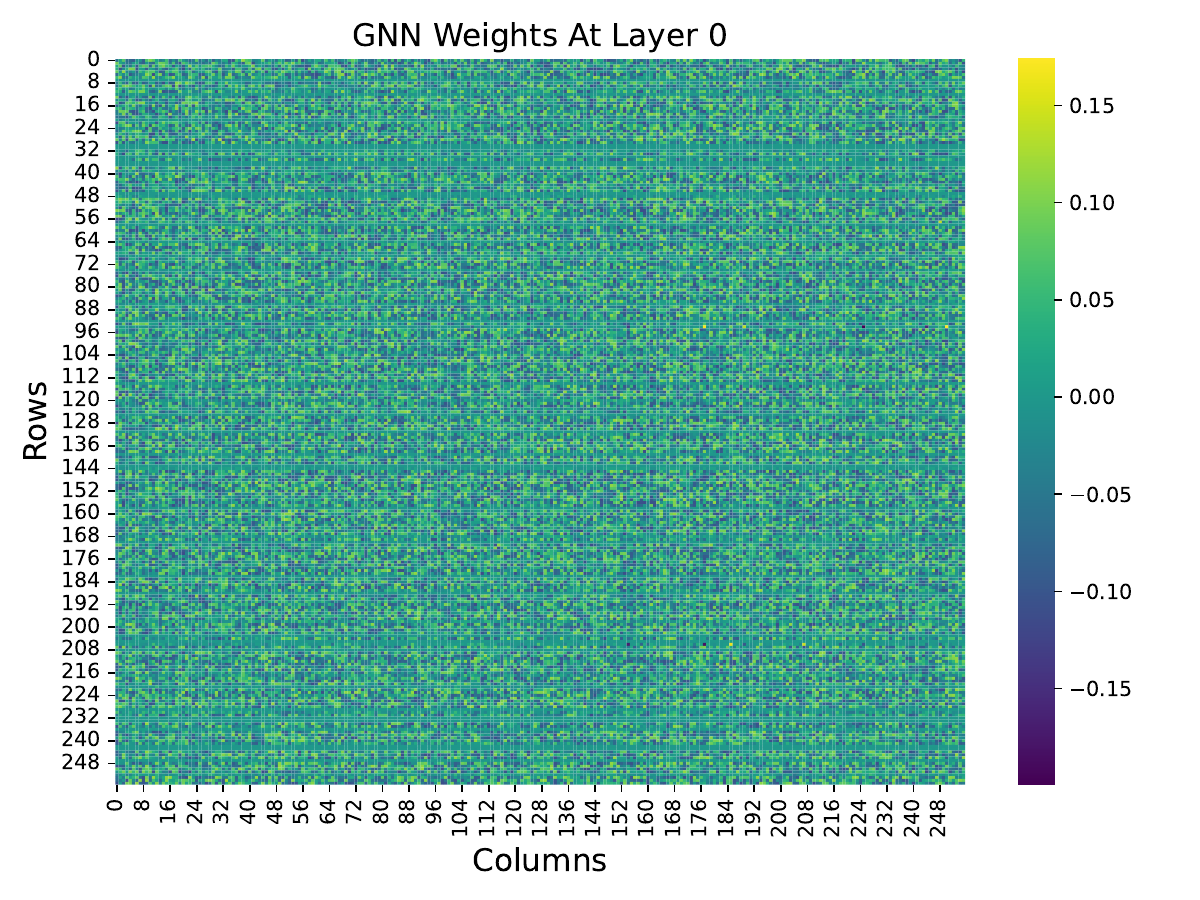} 
        \caption{Layer 1.}
        \label{fig:gcn_ON_weight_matrices_1}
    \end{subfigure}
    \hfill
    \begin{subfigure}[t]{0.48\textwidth}
        \centering
        \includegraphics[width=\textwidth]{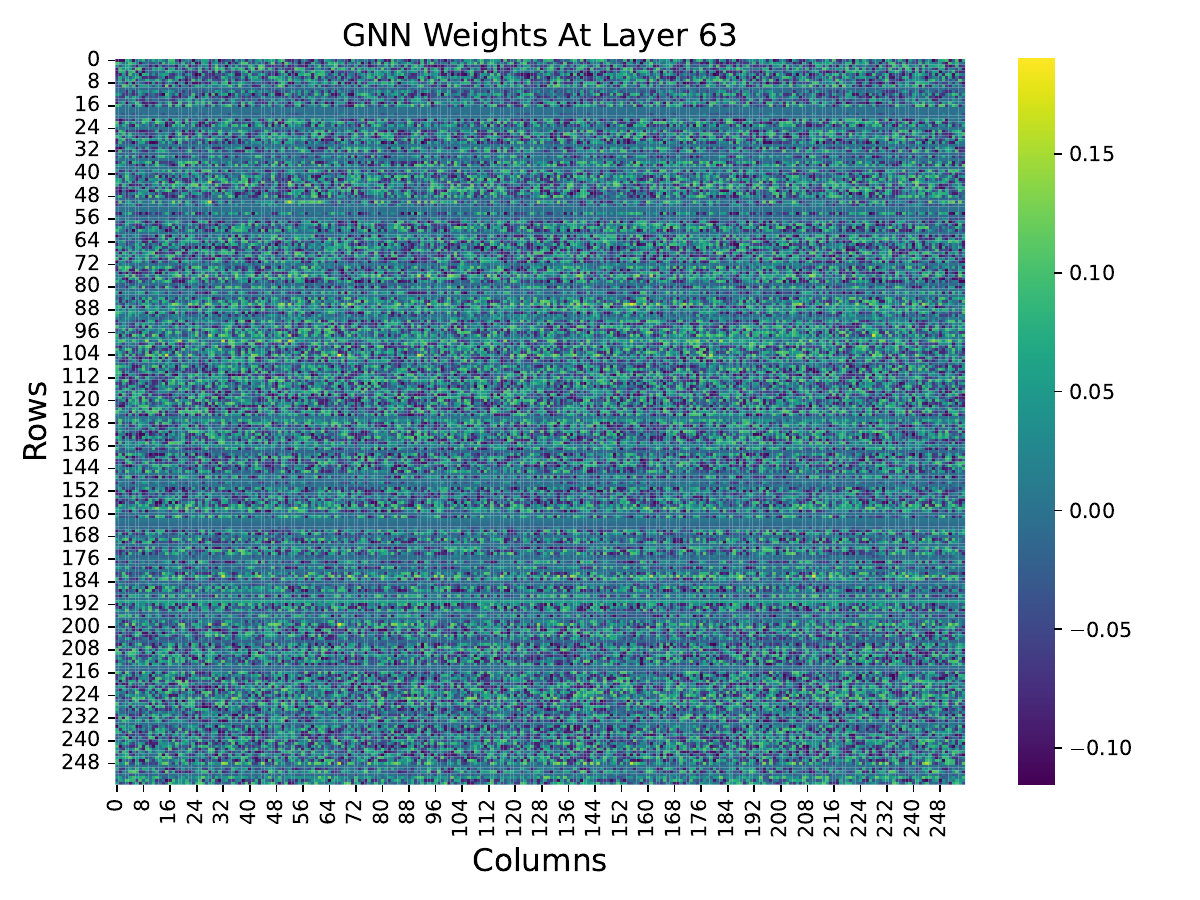} 
        \caption{Layer 64.}
        \label{fig:gcn_ON_weight_matrices_64}
    \end{subfigure}
    \caption{Weight matrices learned in an end-to-end trained GCN with residual connections are shown for layer 1 (a) and layer 64 (b), each sized \(256 \times 256\). Notably, the learned matrices are not diagonal.}
    \label{fig:gcn_weight_matrices}
\end{figure*}

\Cref{theorem:thm} shows that \method mitigates feature collapse especially when compared to GNNs with full random weights of the same depth, we also observe that \emph{in practice \method exhibits significantly higher rank, and therefore less feature collapse, than GNNs with learnable weights}.

These findings align closely with our experimental observations, as shown in \Cref{fig:emb_var}, which compares the rank \(\text{rank}\big(\bold{h}^{(l)}\big)\) across layers for an end-to-end trained GCN, \fixed, \fullfixed, \full, and \method. Both \method and \fixed, which use diagonal weights, exhibit only a slight decrease in the rank of the embedding, in line with  \Cref{theorem:thm}, effectively preventing the feature collapse phenomenon. In contrast, models using full random weights, including End-to-End, \full, and \fullfixed, experience a sharp decline in \(\text{rank}\big(\bold{h}^{(l)}\big)\) to 1, signaling a feature collapse. In addition, we note that, while in end-to-end trained GNNs, the weight matrices, $\mathbf{w}^{(l)}$, could be learned to be diagonal, we do not observe this behavior in practice. Specifically, we illustrate the learned $\mathbf{w}^{(l)}$ at layer 1 and layer 64 for an end-to-end trained GCN (\Cref{fig:gcn_weight_matrices}). In both cases, the learned weight matrices are not diagonal. The results in \Cref{fig:emb_var} and \Cref{fig:gcn_weight_matrices} were conducted on the PubMed dataset using the same GCN architecture with 64 layers ($L = 64$) with residual connections.  

Moreover, as shown in  \Cref{fig:time_analyze_gnn}(c), our  \method, and \fixed consistently maintain stable performance as the number of layers increases, effectively mitigating the challenges posed by feature collapse. In contrast, GNNs with full random weights (\full, \fullfixed, and End-to-End) experience a significant decline in performance with increasing depth due to the loss of embedding diversity. 

The consistency between the theoretical understanding in \Cref{obs:fullmatrix,theorem:thm}  and the empirical results underscores the ability of \method to preserve diversity in embeddings, also in deep architectures. 

\begin{table}[t]
    \centering
    \footnotesize
    \setlength{\tabcolsep}{3pt}
    \begin{tabular}{l  cccccc }
        \toprule
        Method $\downarrow$ / Dataset $\rightarrow$ & 
        \textsc{Cora} &
        \textsc{CiteSeer} &
        \textsc{PubMed} &
        \\
        \midrule  
         End-to-End  & 81.50 $\pm$ 4.8 & {71.10 $\pm$ 0.7} & {79.00 $\pm$ 0.6}  \\ 
         \midrule 
         \identity & 80.36 $\pm$ 0.4 & 69.32 $\pm$ 0.2 & 78.91 $\pm$ 0.3\\ 
         \fixed & 81.32 $\pm$ 0.8 & 70.12 $\pm$ 0.9 & 78.00 $\pm$ 0.9 \\ 
         \fullfixed & 58.70 $\pm$ 1.8 & 55.88 $\pm$ 1.2 & 71.52 $\pm$ 0.6 \\
         \full & 23.29 $\pm$ 9.2 & 45.85 $\pm$ 7.7 & 67.85 $\pm$ 1.7 \\
        \midrule
         \pretrainedours (Ours) & {82.42 $\pm$ 0.3} & {70.75 $\pm$ 0.3} & {78.94 $\pm$ 0.4}  \\ 
        \bottomrule
    \end{tabular}
    \caption{Node classification accuracy comparison (\%)$\uparrow$ of \method, End-to-End, and other baselines using random weights. The results demonstrate that \pretrainedours, employing on-the-fly diagonal random weights in GNN layers, consistently outperforms all baselines using random weights while achieving competitive accuracy and, in some cases, surpassing End-to-End.}  
    \label{tab:GNN_weights_variants}
\end{table}

\begin{table*}[t]
    \centering
    \footnotesize
    \setlength{\tabcolsep}{3pt}    \begin{tabular}{l  ccccc }
        \toprule
        Method $\downarrow$ / Dataset $\rightarrow$ & 
        \textsc{Cora} &
        \textsc{CiteSeer} &
        \textsc{PubMed} &
        \\
        \midrule  
        End-to-End & 81.50 $\pm$ 0.8 & 70.90 $\pm$ 0.7 & 79.00 $\pm$ 0.6  \\ 
        \idenityemb + Random Diagonal & 78.44 $\pm$ 3.0 & OOM & 73.46 $\pm$ 1.1 \\
        \fixedemb + Random Diagonal & 82.82 $\pm$ 1.0 & 71.48 $\pm$ 0.7 & 78.76 $\pm$ 0.6\\
        \dynamicemb + Random Diagonal & 26.30 $\pm$ 6.1 & 25.22 $\pm$ 0.9 & 37.28 $\pm$ 1.4 \\
        \ours + Random Diagonal & {{84.36} $\pm$ 0.3} & {{72.16} $\pm$ 0.6} & {{79.32} $\pm$ 0.3} \\
        \midrule
        Pretrained Embedding + Random Diagonal (\pretrainedours) & {82.42 $\pm$ 0.3} & {70.75 $\pm$ 0.3} & {78.94 $\pm$ 0.4}  \\ 
        \bottomrule

    \end{tabular}
    \caption{Node classification accuracy (\%)$\uparrow$ using a GCN backbone with different embedding strategies. While Learnable embedding performs well across all datasets, it requires backpropagating through the random GNN layers despite only updating the embedding and the classifier. In contrast, using a Pretrained embedding as in \method avoids backpropating through the GNN layers while achieving competitive performance.}
    \label{tab:pretrain}
\end{table*}

\section{Additional Experiments}
\label{app:more_exps}

In this section, we present additional results to further analyze the performance of \textsc{\method}.  

\noindent First, in \Cref{app:GNN_weights_variants}, we expand on the results of \Cref{sec:a3} by  comparing to an additional baseline of \textsc{\method}. 
In \Cref{app:initial_layer}, we study the impact of pretraining the initial embedding.
In \Cref{app:largegraphs_class}, we provide results for \textsc{\method} on the larger ogbn-proteins dataset, comparing its performance against end-to-end trained networks. Finally, we assess the performance of \method across various batch sizes in \Cref{app:batchsize}, demonstrating its adaptability for practical applications.

\subsection{Identity as an Additional Baseline}
\label{app:GNN_weights_variants}

We consider an additional baseline of \textsc{\method}:
\begin{itemize}
    \item \identity: All GNN layer weights, $\mathbf{w}^{(l)}$, are set to identity matrices.
\end{itemize}

\noindent \Cref{tab:GNN_weights_variants} presents the results for the various \textsc{\method} baselines. Among them, \identity shows the lowest performance, indicating that simply passing input representations unchanged, is insufficient to obtain performance comparable to end-to-end training. On the contrary, \pretrainedours outperforms all its baselines that also do not train the GNN weights, and is competitive to end-to-end, underscoring the effectiveness of on-the-fly randomly sampled diagonal matrices in \textsc{\method}.

\subsection{The Importance of Pretraining the Initial Embedding}
\label{app:initial_layer}

In this section, we examine the importance of pretraining the initial embedding on the performance. 
We explore five setups for the embedding layer: (i) the Identity Embedding (\idenityemb), where the initial embedding corresponds to the identity matrix; (ii) Fixed Random Embedding (\fixedemb), which samples a random full-weight matrix initially and keeps it fixed, (iii) \dynamicemb which samples a random full-weight matrix on-the-fly for every forward call (iv) \ours, where the embedding layer weights are learned alongside the classifier, and lastly, (v) pretraining the embedding, as done in our \pretrainedours. All the embedding layers in this section are constructed as an MLP. 

\noindent The results are presented in \Cref{tab:pretrain}. \dynamicemb setup struggles to learn across all datasets, indicating that randomization with on-the-fly sampling in both the initial embedding and the GNN layers hinders effective learning. Similarly, the \idenityemb setup results in performance drops on Cora and PubMed and leads to Out-of-Memory (OOM) issues on CiteSeer, which has more features, underscoring the impracticality of this approach. Thus, we need an initial embedding to map \(c\) input features into a hidden dimension \(d\), allowing all \(\gnnw\) to be diagonal matrices and preventing OOM issues. layers. 
\noindent \fixedemb shows good performance on Cora and CiteSeer but does not show significant gains on PubMed, and suffers from large variance. 
\ours, which can be viewed as the upper bound for \textsc{\method}, performs well across all datasets, even surpassing the end-to-end trained GCN. However, it requires backpropagation through the entire network despite only updating the embedding and classifier, impacting scalability. \pretrainedours offers a more balanced solution, achieving good performance while avoiding backpropagating through the GNN layers.

\begin{figure}[t]
    \centering
    \includegraphics[width=0.45\textwidth]{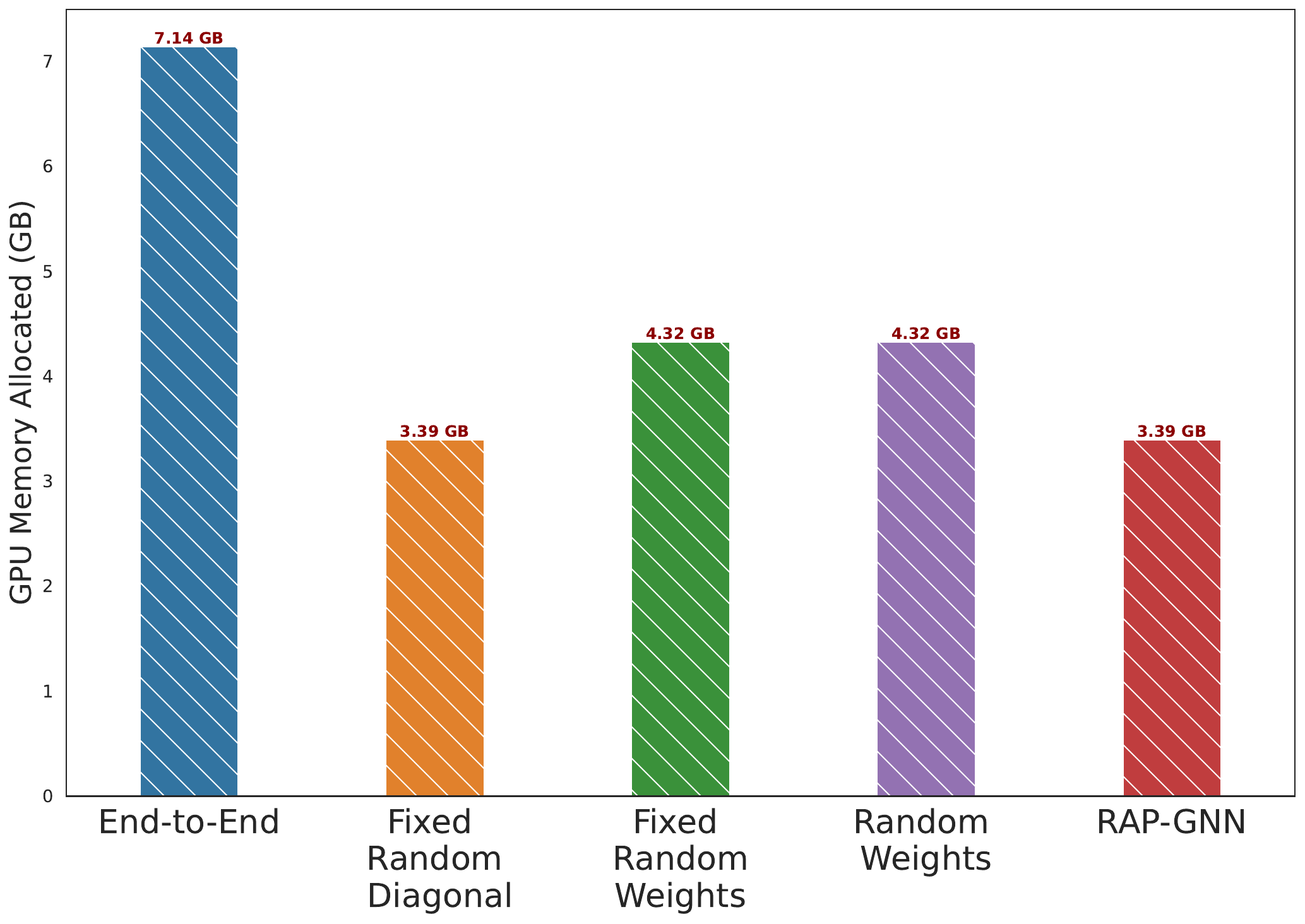} 
    \caption{Comparison of GPU memory consumption on the ogbn-proteins dataset. Our \method significantly reduces GPU memory consumption.}
    \label{fig:gpu_memory_usage_proteins}
\end{figure}

\begin{table*}[ht]
\centering
\footnotesize
\setlength{\tabcolsep}{3pt}
\begin{tabular}{l c cc}
    \toprule
    & \multicolumn{3}{c}{\textsc{ogbn-proteins}}\\
    \cmidrule(l{2pt}r{2pt}){2-4}
         Method $\downarrow$ & ROC-AUC(\%) $\uparrow$ & Train time(ms) $\downarrow$ & Eval. time(ms) $\downarrow$ \\
        \midrule 
        \pretrainedemb & 71.07 $\pm$ 0.36 & 2,803 & 1,540\\
         End-to-End & {77.29} $\pm$ 0.46 & 14,078 &  9,969 \\
         \fixed &  76.84 $\pm$ 0.32 & {7,835}  &  {7,022}   \\
         \fullfixed &  73.71 $\pm$ 0.71 & {9,545}  & {9,920} \\
         \full &  70.46 $\pm$ 0.74 & {9,809}  &  {9,258}  \\
        \midrule
         \pretrainedours (Ours) & 77.01 $\pm$ 0.38 & 7,952 & 7,401\\
    \bottomrule
\end{tabular}
\caption{Node classification ROC-AUC (\%) along with training and evaluation times (ms) on the large ogbn-proteins dataset using a GCN \citep{kipf2016semi} backbone. \textsc{\method} achieve comparable ROC-AUC with end-to-end training, as well as  reduced runtimes.}
\label{tab:ogbn_proteins}
\end{table*}

\begin{table*}[ht]
\centering
\footnotesize
\setlength{\tabcolsep}{3.5pt}
\begin{tabular}{clccc}
\toprule
& & \multicolumn{3}{c}{\textsc{ogbn-products}}\\
    \cmidrule(l{2pt}r{2pt}){3-5}
Backbone & Method & Accuracy (\%) $\uparrow$ & Train time (ms) $\downarrow$ & Eval. time (ms) $\downarrow$ \\
\midrule
\multirow{6}{*}{\rotatebox{90}{GCN}} &
\pretrainedemb & 72.83 $\pm$ 0.40 & 4,722 & 2,861\\
& End-to-End & 75.90 $\pm$ 0.31 & 15,998 & 12,023 \\
& \fixed & 74.93 $\pm$ 0.78 & {9,801} & {9,094} \\
& \fullfixed & 27.72 $\pm$ 0.79 & 13,036 & 12,016\\
& \full & 26.08 $\pm$ 0.82 & 13,164 & 12,165 \\
& \pretrainedours (Ours) & 75.60 $\pm$ 0.88 & {9,861} & {9,124} \\ 
\midrule
\multirow{6}{*}{\rotatebox{90}{GraphSage}} & \pretrainedemb & 74.72 $\pm$ 0.11 & 3,338 & 2,084 \\
& End-to-End & 78.29 $\pm$ 0.16 & 11,865 & 8,548 \\  
& \fixed  & 76.28 $\pm$ 0.39 & {7,748} & {7,083}\\
& \fullfixed  & 27.03 $\pm$ 0.47 & 9,895 & 9,082 \\
& \full  & 26.90 $\pm$ 0.32 & 9,926 & 9,148 \\
& \pretrainedours (Ours) & 77.74 $\pm$ 0.43 & {7,934} & {7,109} \\ 
\bottomrule
\end{tabular}

\caption{Performance Comparison on ogbn-products using GCN \citep{kipf2016semi} and GraphSage \citep{hamilton2017inductive} backbones. Accuracy (\%), Training Time (ms), and Evaluation Time (ms) are reported.}
\label{tab:ogbn-products-full}
\end{table*}

\subsection{Results on Large Scale Graphs} 
\label{app:largegraphs_class}

\noindent {In addition to the accuracy results presented in \Cref{tab:ogbn-products}, we further evaluate the performance of \method on the ogbn-proteins and ogbn-products datasets, including running time and GPU memory consumption. All experiments were conducted on an NVIDIA RTX 4090.}

\paragraph{ogbn-proteins.}
To evaluate the scalability of \textsc{\method}, we conduct experiments on the large-scale ogbn-proteins dataset. 
We evaluate \textsc{\method} and all baselines using a GCN backbone with residual connections. The pretrained embedding layer in \textsc{\method} and its baselines that also sample random weights consists of exactly one single GNN layer. 
Further details on the implementation and hyperparameter search can be found in \Cref{app:hyperparameter_search}.

\noindent As shown in \Cref{tab:ogbn_proteins}, \method achieves performance comparable to end-to-end while requiring significantly less time per epoch for both training and evaluation, as well as using less than half of the memory of an end-to-end GCN (\Cref{fig:gpu_memory_usage_proteins}). Although \fixed also achieves efficiency in time and GPU memory consumption, its performance is slightly lower than that of end-to-end and our \method. On the other hand, \full and \fullfixed exhibit poor performance in both efficiency and efficacy.

\paragraph{ogbn-products.}
We further evaluate the performance of \method on the ogbn-products dataset with the GraphSage backbone \citep{hamilton2017inductive}, as well as the running time of \method and all baselines.
\noindent The results in \Cref{tab:ogbn-products-full} demonstrate that \pretrainedours offers a balanced approach between accuracy and computational efficiency when compared to End-to-End in both backbones. Specifically, we note that on both GCN and GraphSage backbones, our \method achieves comparable accuracy to its corresponding  end-to-end trained counterpart, while requiring significantly reduced training and evaluation times. In contrast, other baselines using random weights (\full, \fullfixed, and \fixed) show a drop in performance, although they  demonstrate gains in efficiency.

{\subsection{Effect of Batch Size on \method Performance}  
\label{app:batchsize}
Since each \(\mathbf{w}^{(l)}\) is sampled anew for every forward pass — remaining consistent across all graphs within the same batch but varying between batches — it is important to evaluate the impact of batch size. To this end, we conducted additional experiments using the PROTEINS dataset. The results, summarized in \Cref{tab:batchsize_effect}, show that batch size has minimal influence on the performance of \method. Specifically, \method achieves consistent accuracy across varying batch sizes, with only minor variations observed. These findings demonstrate the robustness of \method to different batch size settings, underscoring its flexibility in practical applications. 

\begin{table}[t]
    \centering
    \footnotesize
    \setlength{\tabcolsep}{3pt}
    \begin{tabular}{lccc}
        \toprule
& \multicolumn{3}{c}{\textsc{PROTEINS}}\\
    \cmidrule(l{2pt}r{2pt}){2-4}
        Method & Batchsize=256 & Batchsize=512 & Batchsize=1024 \\
        \midrule
        RAP-GNN (Ours) & 75.1 $\pm$ 3.3 & 75.9 $\pm$ 3.8 & 75.0 $\pm$ 3.1 \\
        \bottomrule
    \end{tabular}
    \caption{Performance of RAP-GNN with varying batch sizes on TUDataset (PROTEINS) using GIN backbone \citep{xu2018how}.}
    \label{tab:batchsize_effect}
\end{table}
}

\begin{table}[t]
\centering
\footnotesize
\setlength{\tabcolsep}{3.5pt}
\begin{tabular}{l c}
    \toprule
        \multirow{2}*{Method $\downarrow$} & \textsc{molbace} \\
        & \textsc{ROC-AUC $\uparrow$} \\
        \midrule  
        \pretrainedemb & 65.64 $\pm$ 3.57 \\
        End-to-End & 72.97 $\pm$ 4.00 \\
        \fixed & 72.68 $\pm$ 3.10 \\
        \fullfixed & 56.80 $\pm$ 5.55 \\
        \full & 55.23 $\pm$ 6.21 \\
        \midrule
        \pretrainedours (Ours) & 73.42 $\pm$ 2.59 \\
        \bottomrule
\end{tabular}
\caption{Graph classification performance (\%)$\uparrow$ of \textsc{\method} on the molbace dataset.}
\label{tab:molbace}
\end{table}

\subsection{Results on ogbg-molbace}
\label{app:molbace}
In addition to the results on the OGB datasets presented in \Cref{tab:no_molbace}, we now include results for the ogbg-molbace dataset in \Cref{tab:molbace}. Similar to the patterns observed in other datasets, \method outperforms all evaluated baselines.

\section{Dataset and Experiment Details}
\label{app:hyperparameter_search}

\subsection{Dataset Statistics}
In this section, we provide a summary of all the datasets used in this paper, as detailed in \Cref{tab:data_stats}. The datasets contain a variety of types, graph tasks, and sizes, ensuring a fair comparison and comprehensive evaluation across experiments.

\begin{table*}[t]
\footnotesize
\setlength{\tabcolsep}{3pt}
\centering
\begin{tabular}{l l ccccc}
\toprule
& Dataset & Task & \#Graphs & Avg. \#Nodes & Avg. \#Edges & \#Classes \\
\midrule
\multirow{3}{*}{\textbf{Planetoid}} & \textsc{Cora}     & N  & 1     & 2,708   & 10,556   & 7 \\
&\textsc{CiteSeer} & N  & 1     & 3,327   & 9,104    & 6 \\
&\textsc{PubMed}   & N  & 1     & 19,717  & 88,648   & 3 \\
\midrule
\multirow{3}{*}{\textbf{TUDatasets}} & \textsc{MUTAG}    & G  & 188   & 17.9    & 19.7     & 2 \\
& \textsc{PTC}      & G  & 344   & 14.29   & 14.69    & 2 \\
& \textsc{PROTEINS} & G  & 1,113 & 39.1    & 72.8     & 2 \\
\midrule
\multirow{5}{*}{\textbf{OGB}} & \textsc{molesol}  & G  & 1,128 & 13.3    & 13.7    & - \\
& \textsc{moltox21} & G  & 7,831 & 18.6    & 19.3    & - \\
& \textsc{molhiv}   & G  & 41,127& 25.5    & 27.5     & 2 \\
& \textsc{molbace}  & G  & 1,513 & 34.1    & 36.9     & 2 \\
& \textsc{ogbn-proteins}   & N  & 1 & 132,534    & 39,561,252    & 112 \\
& \textsc{ogbn-products}   & N  & 1 & 2,449,029    & 61,859,140 & 47 \\
\bottomrule
\end{tabular}
\caption{Overview of the graph learning datasets. N denotes a Node-level task, and G denotes a Graph-level task.}
\label{tab:data_stats}
\end{table*}

\subsection{Experimental Details} 

\begin{table*}[t]
    \centering
    
    \small
    \begin{tabular}{lcccccc} 
    \toprule
        Hyperparameter & \textsc{molesol} & \textsc{moltox21} & \textsc{molhiv} & \textsc{molbace} & \textsc{ogbn-proteins} & \textsc{ogbn-products} \\
        \midrule 
        \#\textsc{layer} in $\pretraininitlayer$ & 1 & 1 & 1 & 1 & 1 & 1 \\
        \#\textsc{layer} in $\classifier$ & 1 & 1 & 1 & 1 & $\{2, 3, 4\}$ & 1 \\
        L & $\{1, 2, 3\}$ & $\{2, 3, 4\}$ & $\{1, 2, 4, 6, 8\}$ & $\{2, 4, 6\}$ & $\{3, 4, 6, 8\}$ & $[2,3,4]$ \\
        \textsc{lr} & $[0.001, 0.05]$ & $[0.005, 0.01]$ & $[0.001, 0.05]$ & $[0.0005, 0.01]$ & $[0.007, 0.1]$ & $[0.003, 0.03]$ \\
        \textsc{hidden dim.} & $\{16, 32\}$ & $\{8, 16\}$ & $\{32, 64, 128, 256\}$ & $\{32, 64, 128\}$ & $\{240, 360\}$ & $[256, 300]$ \\
        \textsc{\#epochs} & 350 & $\{350, 500\}$ & $\{350, 500, 1000\}$ & 350 & $[500, 1000]$ & $[20,30]$ \\
        \textsc{batch size} & $\{256, 512\}$ & 512 & -- & 512 & -- & -- \\
        \textsc{dropout} & $[0, 0.6]$ & $[0, 0.5]$ & $[0, 0.5, 0.6]$ & $[0.4, 0.6]$ & $[0.2, 0.6]$ & $[0.15,0.3]$ \\
        \bottomrule
    \end{tabular}
    \caption{Hyperparameter search for the datasets in the OGB collection: molesol, moltox21, molhiv, molbace, {ogbn-proteins}, and ogbn-products.}
    \label{tab:hyperparameters_ogb}
\end{table*}

\begin{table}[t]
    \centering
    \footnotesize
    \setlength{\tabcolsep}{2.8pt}
    \begin{tabular}{lccc} 
    \toprule
        Hyperparameter & \textsc{Cora} & \textsc{CiteSeer} & \textsc{PubMed} \\
        \midrule 
        \#\textsc{layer} in $\pretraininitlayer$ & $\{1, 2\}$ & $\{1, 2, 3\}$ & $\{1, 2, 3\}$ \\
        \#\textsc{layer} in $\classifier$ & 1 & 1 & 1 \\
        L & $\{2, 4, 8, 16\}$ & $\{2, 4, 6\}$ & $\{4, 6, 8, 12\}$ \\
        \textsc{lr} & $[0.001, 0.01]$ & $[0.001, 0.01]$ & $[0.001, 0.01]$ \\
        \textsc{hidden dim.} & $\{16, 32\}$ & $\{16, 32, 64, 128\}$ & $\{16, 64, 128, 256\}$ \\
        \textsc{\#epochs} & 700 & 1000 & 1000 \\
        \textsc{batch size} & -- & -- & -- \\
        \textsc{dropout} & $[0, 0.5]$ & $[0, 0.5]$ & $[0, 0.5]$ \\
        \bottomrule
    \end{tabular}
    \caption{Hyperparameter search for the Planetoid datasets. 
    }
    \label{tab:hyperparameters_planetoid}
\end{table}

\begin{table}[t]
    \centering
    \footnotesize
    \setlength{\tabcolsep}{2.8pt}
    \begin{tabular}{lccc} 
    \toprule
        Hyperparameter & \textsc{MUTAG} & \textsc{PTC} & \textsc{PROTEINS} \\
        \midrule 
        \#\textsc{layer} in $\pretraininitlayer$ & $\{1, 2, 3\}$ & $\{1, 2, 3\}$ & $\{1, 2, 3\}$ \\
        \#\textsc{layer} in $\classifier$ & 2 & 2 & 1 \\
        L & $\{1, 2, 3\}$ & $\{2, 3, 4, 6\}$ & $\{1, 2, 3, 4\}$ \\
        \textsc{lr} & $[0.007, 0.05]$ & $[0.007, 0.05]$ & $[0.007, 0.05]$ \\
        \textsc{hidden dim.} & $\{16, 64, 128\}$ & $\{16, 64, 128\}$ & $\{16, 32, 64\}$ \\
        \textsc{\#epochs} & 1000 & 1000 & 1000 \\
        \textsc{batch size} & $\{64, 128, 512\}$ & $\{64, 128, 512\}$ & $\{128, 256, 512\}$ \\
        \textsc{dropout} & $[0, 0.35]$ & $[0, 0.35]$ & $[0, 0.35]$ \\
        \bottomrule
    \end{tabular}
    \caption{Hyperparameter search for the TUDatasets. 
    }
\label{tab:hyperparameters_tu}
\end{table}

Our experiments were conducted using the PyTorch \citep{paszke2019pytorch} and PyTorch Geometric \citep{fey2019fast} frameworks, utilizing Weight and Biases \citep{wandb} for hyperparameter sweeps. In this section, we provide details on our specific implementation of the experiments. 

\noindent \Cref{tab:hyperparameters_planetoid,tab:hyperparameters_tu,tab:hyperparameters_ogb} outline the hyperparameter search space for all datasets divided by their collections. Specifically, \Cref{tab:hyperparameters_planetoid} includes all datasets from the Planetoid collection, \Cref{tab:hyperparameters_tu} presents datasets from the TUDatasets collection, and \Cref{tab:hyperparameters_ogb} outlines the OGB collection. Note that all \(\mathbf{w}^{(l)}\) matrices must be square, which requires the input and output dimensions of all hidden layers in \(\gnnlayerl\) to be the same, corresponding to the \textsc{hidden dim}. We use \(L\) to denote the number of hidden GNN layers. Square brackets \([]\) indicate ranges, and curly braces \(\{\}\) represent sets for all tables.

\paragraph{Node Classification Tasks.} For the Cora, CiteSeer, and PubMed datasets, all networks use a GCN backbone. The end-to-end trained networks implementation follows the PyTorch Geometric example, incorporating residual connections \citep{avelar2019discrete}. All baselines of \textsc{\method} use MLP-based embeddings as the initial node embeddings. The number of MLP layers for the embedding is treated as a hyperparameter, while the hidden and output dimensions of the MLP are equal to $d$. All hyperparameter search details are in \Cref{tab:hyperparameters_planetoid}.

\noindent For the ogbn-proteins dataset, the implementation is based on \citet{luo2024classic} with a GCN backbone. The embedding is a single GCN layer. The evaluation metric for the ogbn-proteins dataset follows \citet{hu2020ogb} and uses the ROC-AUC for binary classification. Details of all hyperparameters are given in \Cref{tab:hyperparameters_ogb}.

\noindent For the ogbn-products dataset, we construct \method using the implementation of a GCN or GraphSage backbone provided by \citet{wang2019dgl}. The embedding consists of a single GCN or GraphSage layer, matching the GNN layers used in the model. The evaluation metric for this dataset aligns with the multi-class classification approach outlined in \citet{hu2020ogb}. Detailed hyperparameter settings are provided in \Cref{tab:hyperparameters_ogb}.

\noindent In all node classification tasks, batching is not required for either training or evaluation; therefore, the batch size rows in \Cref{tab:hyperparameters_planetoid,tab:hyperparameters_ogb} are left blank.
For all considered experiments in all node classification tasks, we show the mean $\pm$ std. of 5 runs with different random seeds. 

\paragraph{Graph Classification Tasks.} For the TUDatasets, we use the GIN baseline (employing 1-hidden-layer MLP inside) implementation from the PyTorch Geometric example. All baselines of \textsc{\method} use MLP-based initial node embeddings, with the number of layers treated as a hyperparameter, and hidden and output dimensions of the MLP equal to $d$. Additionally, we follow the procedure in \cite{xu2018how}, namely, we perform 10-fold cross-validation and report the validation mean and standard deviation at the epoch that achieved the maximum averaged validation accuracy. Details of all hyperparameters are provided in \Cref{tab:hyperparameters_tu}.

\noindent For the OGB datasets, we adopt the GIN network for training and evaluation, following the example code from \citet{hu2020ogb}. The embedding $\pretraininitlayer$ is identical to all other GNN layers, incorporating a 2-hidden-layer MLP with hidden and output dimensions equal to $d$ within each GIN layer. The four datasets used for evaluation are molesol, moltox21, molbace, and molhiv from \citet{hu2020ogb}.  Furthermore, for all considered experiments under node classification using OGB datasets, we show the mean $\pm$ std. of 5 runs with different random seeds. More details of the hyperparameter search are in \Cref{tab:hyperparameters_ogb}.

\end{document}